\documentclass[letterpaper]{article} 
\usepackage{aaai24}  
\usepackage{times}  
\usepackage{helvet}  
\usepackage{courier}  
\usepackage[hyphens]{url}  
\usepackage{graphicx} 
\usepackage{natbib}  
\usepackage{caption} 
\usepackage{algorithm}
\usepackage{algorithmic}

\usepackage{newfloat}
\usepackage{listings}

\usepackage{amsmath,epstopdf,amssymb,color,url,multirow,multicol,verbatim,threeparttable,diagbox,makecell,booktabs}

\newcommand{\tf}{}
\newcommand{\tft}{\textbf}

\newcommand{\etal}{\textit{et al.}}

\newcommand{\cred}{\color{red}}

\DeclareCaptionStyle{ruled}{labelfont=normalfont,labelsep=colon,strut=off} 
\floatstyle{ruled}
\newfloat{listing}{tb}{lst}{}
\floatname{listing}{Listing}
\title{Frequency-Aware Deepfake Detection: Improving Generalizability through Frequency Space Learning}
\author{Chuangchuang Tan$^{1,2}$, Yao Zhao$^{1,2}$\thanks{Corresponding author}, Shikui Wei$^{1,2}$, Guanghua Gu$^{3,4}$, Ping Liu$^{5}$, Yunchao Wei$^{1,2}$ \\
{\small $^1$Institute of Information Science, Beijing Jiaotong University}\\
{\small $^2$Beijing Key Laboratory of Advanced Information Science and Network Technology} \\
{\small $^3$School of Information Science and Engineering, Yanshan University}\\
{\small $^4$Hebei Key Laboratory of Information Transmission and Signal Processing}\\
{\small $^5$Center for Frontier AI Research, IHPC, A*STAR, Singapore}\\
{\tt\small \{tanchuangchuang, yzhao, shkwei\}@bjtu.edu.cn, guguanghua@ysu.edu.cn, pino.pingliu@gmail.com, wychao1987@gmail.com}
}

\usepackage{bibentry}

\begin{document}

\maketitle

\begin{abstract}
This research addresses the challenge of developing a universal deepfake detector that can effectively identify unseen deepfake images despite limited training data.  Existing frequency-based paradigms have relied on frequency-level artifacts introduced during the up-sampling in GAN pipelines to detect forgeries. However, the rapid advancements in synthesis technology have led to specific artifacts for each generation model. 
Consequently, these detectors have exhibited a lack of proficiency in learning the frequency domain and tend to overfit to the artifacts present in the training data, leading to suboptimal performance on unseen sources. To address this issue, we introduce a novel frequency-aware approach called FreqNet, centered around frequency domain learning, specifically designed to enhance the generalizability of deepfake detectors. Our method forces the detector to continuously focus on high-frequency information, exploiting high-frequency representation of features across spatial and channel dimensions. Additionally, we incorporate a straightforward frequency domain learning module to learn source-agnostic features. It involves convolutional layers applied to both the phase spectrum and amplitude spectrum between the Fast Fourier Transform (FFT) and Inverse Fast Fourier Transform (iFFT). Extensive experimentation involving 17 GANs demonstrates the effectiveness of our proposed method, showcasing state-of-the-art performance (+9.8\%) while requiring fewer parameters. The code is available at {\cred \url{https://github.com/chuangchuangtan/FreqNet-DeepfakeDetection}}.

\end{abstract}

\section{Introduction}

The proliferation of Generative Adversarial Networks (GANs) \cite{goodfellow2014generative, karras2018progressive, karras2019style} has significantly simplified the generation of lifelike synthetic images, resulting in an alarming surge in the prevalence of forgeries that are virtually indistinguishable from authentic images to the human visual system. This escalating trend poses potential, unpredictable societal repercussions. 
In response, a multitude of deepfake detection mechanisms have been conceived \cite{Frank,li2021frequency},
with specific emphasis on detecting facial forgeries. However, the majority of existing forgery detection techniques suffer from a fundamental limitation: they are constrained to the same domain during both their training and evaluation phases. This limitation severely hampers their ability to generalize effectively to unseen domains, such as those involving unfamiliar generation models or novel categories.

\begin{figure}[t]
  \centering
   \includegraphics[scale=0.52]{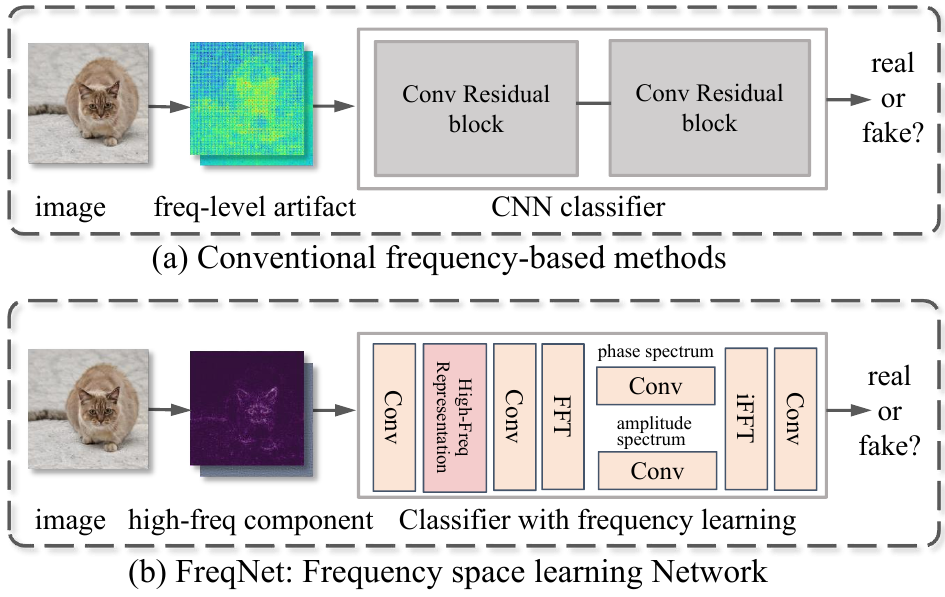}
   \caption{{Frequency space learning network.} 
 (a) The traditional studies are usually limited to developing frequency-level artifacts. (b) Distinguishing itself from prior frequency-based research, our approach shifts its focus to the frequency-related attributes of the features within the detector. This novel perspective includes a continuous emphasis on high-frequency details within the classifier, which capitalizes on the enriched depiction of high-frequency feature map components spanning both spatial and channel dimensions. Additionally, our strategy introduces a trainable layer embedded within the frequency domain, facilitating the acquisition of source-agnostic features.
}
   \label{fig:fig0}
\end{figure}

The pursuit of Generalizable Deepfake Detection strives to create a universal detector capable of effectively identifying deepfake images even when faced with limited training data, a necessity given the continual emergence of increasingly sophisticated synthesis technologies. 
Recently, prior investigations \cite{Frank, Durall} have substantiated the efficacy of frequency artifacts in the realm of deepfake detection, notably in the context of facial detection \cite{qian2020thinking, luo2021generalizing}. 
The findings of \cite{Frank, Durall} have unveiled the presence of significant artifacts within the frequency domain stemming from the upsampling operations within GAN architectures. Consequently, this revelation has spurred the development of numerous frequency-based methods aimed at detecting images with pronounced frequency-related characteristics. 

\begin{figure}[t]
  \centering
   \includegraphics[scale=0.40]{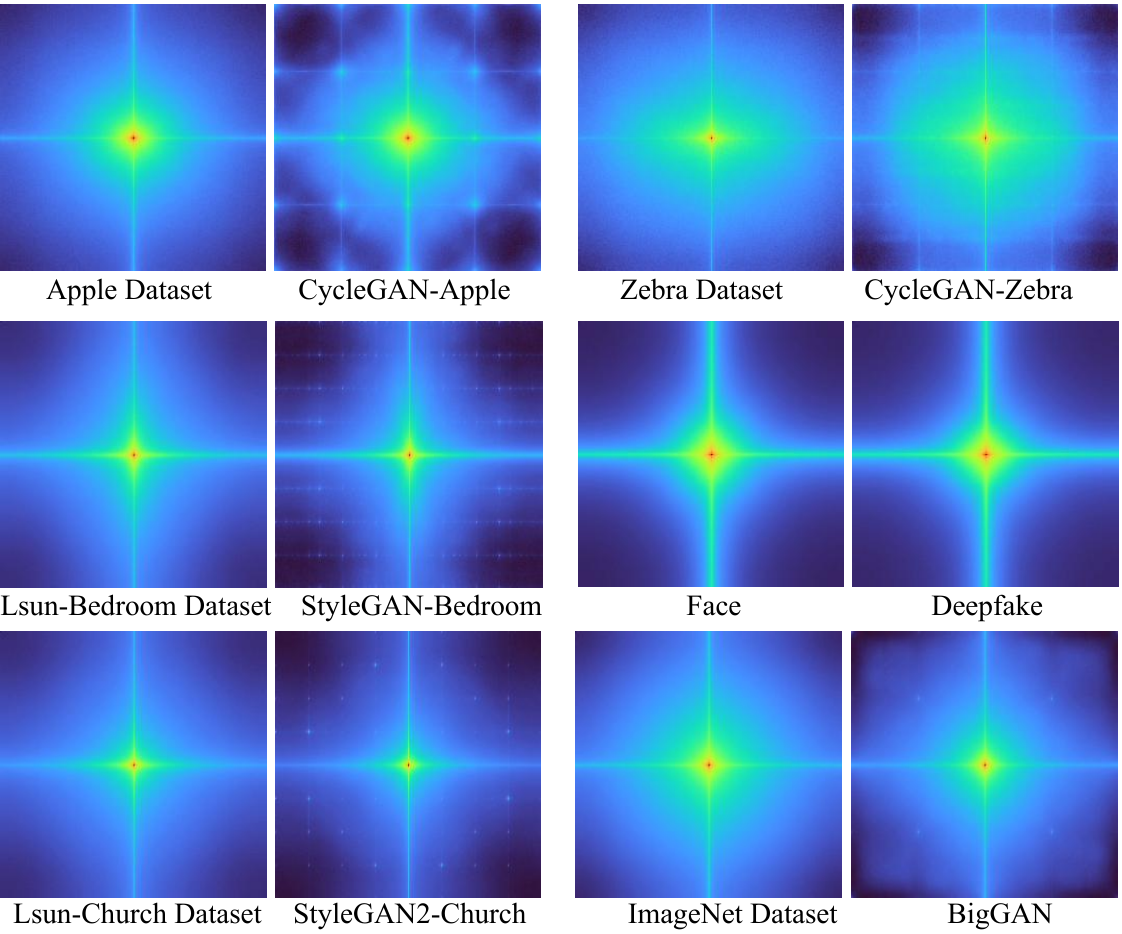}
   \caption{{Frequency analysis on various sources.} This mean FFT spectrum computation involves averaging over 2,000 images, following the methodology detailed in \cite{Frank}. }
   \label{fig:fig1}
\end{figure}

Nonetheless, owing to the extraordinary advancements in synthesis technology, an increasing array of distinctive frequency-level artifact representations have emerged. 
In Figure \ref{fig:fig1}, we present the mean Fast Fourier Transform (FFT) \cite{cooley1969fast} spectrum of images sampled from various sources. 
This mean spectrum computation involves averaging over 2,000 images, following the methodology detailed in \cite{Frank}. 
Notably, the results exhibit discernible differences in artifact characteristics across different GANs, further accentuated by variations within the same GAN architecture when training on dissimilar datasets. 
The frequency attributes of images indeed possess the capacity to unveil distinctions between real and generated images. However, they exhibit limitations in terms of generalization across a diverse range of sources. The CNN classifier, when trained on a specific source (e.g., StyleGAN), tends to overfit to the particular pattern present within the training data. As a result, this classifier often falters when faced with unseen synthesis models such as CycleGAN and BigGAN.

To surmount this challenge, the primary approach entails the development of a robust classifier specifically designed for frequency representations. 
\cite{jeong2022frepgan}, in addressing this issue, chooses to disregard frequency-level artifacts in images by devising a frequency-level perturbation generator. However, this solution introduces complexity and incurs considerable computational expenses. 
In the realm of deepfake detection, it is imperative to factor in the notion of the detector learning within the frequency domain. We deliberately refrain from directly utilizing the frequency information as the artifact representation to train a CNN classifier. Instead, we strategically compel the detector to acquire its understanding within the frequency space. This nuanced strategy holds the key to achieving a more generalizable deepfake detection framework.

Drawing upon intuitive insights, we introduce a novel and lightweight approach named FreqNet, which integrates frequency domain learning into a lightweight CNN classifier, aimed at enhancing the generalization capabilities of the detector. Diverging from existing frequency-based studies that predominantly focus on the frequency domain of images, the principal innovation of our FreqNet method resides in the simultaneous development of both frequency-domain information derived from images and the features extracted by CNN model. This distinctive dual approach empowers the detector to acquire proficiency in the frequency domain, thereby diminishing its reliance on the specific frequency patterns present in the training source.

Specifically,  our approach introduces two critical modules: the high-frequency representation and the frequency convolutional layer, each meticulously designed to facilitate frequency space learning. The first module serves to compel the detector to consistently prioritize high-frequency information, augmenting its sensitivity to significant details. Additionally, to capture broader forgery indicators within the frequency domain, we incorporate a frequency convolutional layer, which effectively diminishes the reliance on source-specific characteristics. By virtue of this frequency domain-based learning strategy, our proposed FreqNet remarkably extends its generalizability to previously unseen sources with few parameters.

To comprehensively assess the extent of its generalization capabilities, we conduct  extensive simulations using an extensive image database generated by 17 distinct models~\footnote{ProGAN, StyleGAN, StyleGAN2, BigGAN, CycleGAN, StarGAN, GauGAN, Deepfake, AttGAN, BEGAN, CramerGAN, InfoMaxGAN, MMDGAN, RelGAN, S3GAN, SNGAN, STGAN}.
Despite its modest scale, FreqNet, boasting 1.9 million parameters, significantly outperforms the current state-of-the-art model boasting 304 million parameters, demonstrating a remarkable improvement of 9.8\%.

Our paper makes the following contributions:
\begin{itemize}
\item We present a novel frequency space learning network, FreqNet, to achieve generalizable deepfake detection. Our approach strategically incorporates frequency domain learning within a CNN classifier, resulting in a significant enhancement of the detector's ability to generalize across diverse scenarios.

\item We utilize convolutional layers on both the phase spectrum and the amplitude spectrum as a deliberate strategy to capture broader forgery indicators within the frequency domain. This allows us to enhance the detector's capability to identify a broader range of artifacts. 

\item The proposed lightweight FreqNet, consisting of a mere 1.9 million parameters, impressively outperforms the current state-of-the-art model featuring 304 million parameters, attributed mainly to its utilization of frequency domain learning. 

\end{itemize}


\section{Related Work}
In this section, we present a concise survey of deepfake detection methodologies, categorizing them into two primary classes: image-based detection and frequency-based detection.

\subsection{Image-based Deepfake Detection}
There has been a significant effort in the field of forgery detection, with many studies focusing on leveraging spatial information from images.
 Rossler \etal \cite{Deepfake} utilize images to train the Xception \cite{chollet2017xception} to achieve fake face image detection. 
Other image-based detection methods have developed specific artifact detection in distinct facial regions, such as eyes\cite{li2018ictu} and lips \cite{haliassos2021lips}.
Chai \etal \cite{chai2020makes} adopt limited receptive fields to identify patches that render images detectable, highlighting the importance of specific local features.  
With the emergence of deepfake technology, efforts are being made to enhance the generalization ability of detectors, particularly for unseen data.  Yu \etal \cite{yu2020mining} introduce artifacts from the camera imaging process. Furthermore, various methods \cite{wang2020cnn, wang2021representative,chen2022self,cao2022end, he2021beyond, shiohara2022detecting} aim to enrich the diversity of training data, employing techniques such as data augmentation, adversarial training, reconstruction, and blending images. CDDB \cite{li2023continual} adopts incremental learning to achieve continual deepfake detection.  Notably, recent works by Ojha \etal \cite{ojha2023towards} and Tan  \etal \cite{Tan2023CVPR} employ the feature map and gradients as the general representation, respectively.

\subsection{Frequency-based Deepfake Detection}
The research by \cite{Frank, Durall}, which highlights the effectiveness of frequency artifacts in the domain of deepfake detection, has significantly influenced subsequent studies. These findings have led many image forgery detectors to shift their attention toward capturing unique patterns within the frequency domain. 
The work by Masi et al. \cite{masi2020two} stands out as it meticulously investigates artifacts present in both the color space and the frequency domains, while \(F^3\)-Net \cite {qian2020thinking} suggests using the discrepancy of frequency statistics between real and forged images as a means to differentiate face image manipulations. 
An adaptive frequency features learning is designed by FDFL \cite{li2021frequency} to mine subtle artifacts from the frequency domain, enabling the detection of forged images. 
In parallel, Luo \etal \cite{luo2021generalizing} improve the generalization performance through the integration of multiple high-frequency features, thereby fortifying the robustness of the forgery detection process. 
Moreover, ADD \cite{woo2022add} incorporates two meticulously designed distillation modules to emphasize the significance of frequency information, incorporating frequency attention distillation and multi-view attention distillation. 
Recently, frequency-based studies have been proposed for generalized detection.
BiHPF \cite{jeong2022bihpf} emphasizes amplifying artifact magnitudes through the utilization of dual high-pass filters.  
The FreGAN model, introduced by \cite{jeong2022frepgan} ingeniously mitigates the impact of frequency-level artifacts through the deployment of frequency-level perturbation maps. \cite{wang2023dynamic} introduces dynamic graph learning to exploit the relation-aware features in spatial and frequency domains.

\begin{figure*}[t]
  \centering
   \includegraphics[scale=0.530]{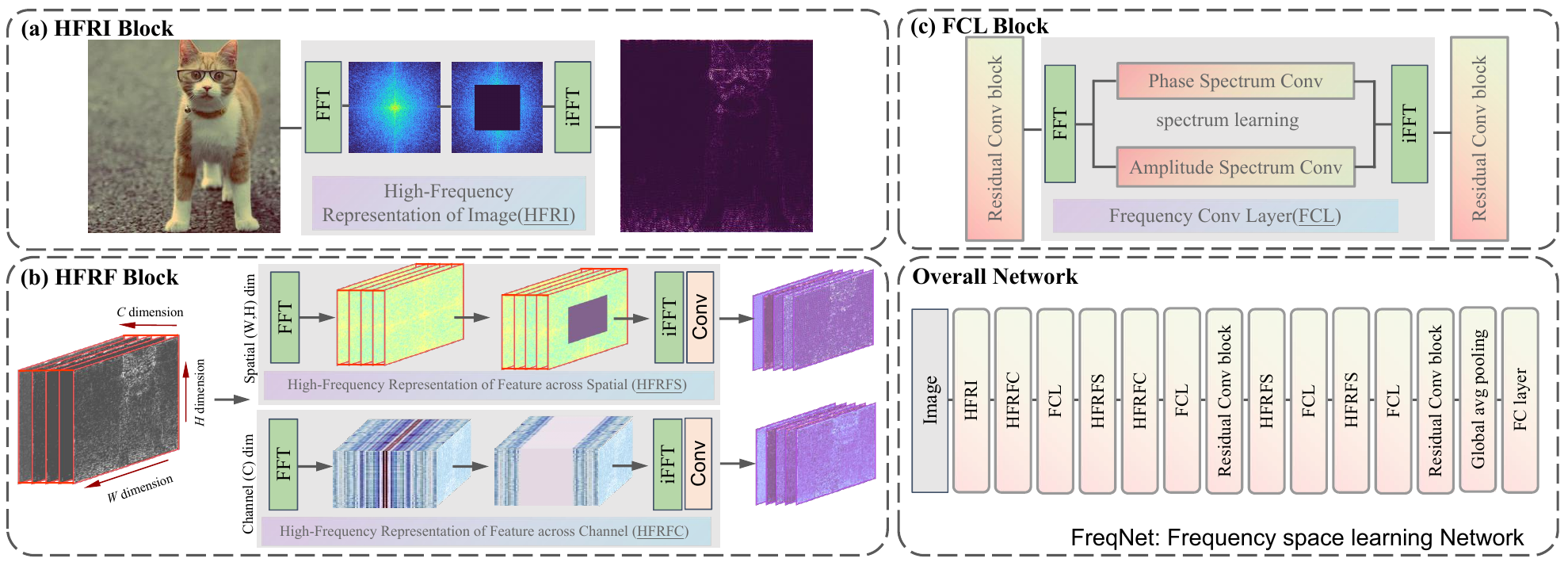}
   \caption{{Architecture of FreqNet for generalizable deepfake detection.} To augment the capacity for generalization, our FreqNet  focuses on the enhancement of frequency spectrum information, prioritizing frequency domain learning within the classifier,  consisting of (a) High-Frequency
Representation of Image(HFRI), (b) High-Frequency Representation of Feature(HFRF), and (c) Frequency Conv Layer(FCL).}
   \label{fig:overall}
\end{figure*}

\section{Methodology}

In this section, we present our FreqNet technique, a universal deepfake detection method for generalizable deepfake detection. 
We illustrate the overall architecture of FreqNet  in Figure \ref{fig:overall}, leveraging frequency domain learning to mitigate source-specific dependencies. 

\subsection{Problem Definition} 
Our primary focus lies within the realm of Generalizable Deepfake Detection. Our objective is to construct a universal detector capable of accurately identifying deepfake images even when confronted with constrained training sources. In this context, let us consider a real-world image scenario denoted as $X$ sampled from $n$ different sources:
\begin{equation}
\begin{split}
    &X = \{  X_{1}, X_{2}, ... ,X_{i}, ..., X_{n} \}, \\ 
    &X_{i} =  \{ x_{j}^{i}, y_{j}\}_{j=1}^{N_{i}},  
 \end{split}
  \label{eq:eq1}
\end{equation}
where $N_i$ represents the number of images originating from the $i$th source $X_{i}$, $x_{j}^{i}$ is the $j$th image of $X_{i}$. Each image is labeled with $y$, indicating whether it belongs to the category of "real"$(y=0)$ or "fake" $(y=1)$. Here we train a binary classifier \(D(\cdot)\), utilizing the training source $X_{i}$:
 \begin{equation}
 D^{{i}} = \mathop{\arg\min}_{\theta} \ l(D( X_{i}; \theta),\  y),
  \label{eq:eq3}
\end{equation}
where $l()$ denote the loss function. 
Our overarching goal is to design a detector that is trained on the data originating from $X_i$, but demonstrates strong performance when confronted with images coming from previously unseen sources, denoted as $X_{t}$. This generalizability across unseen sources is a crucial objective of our detector.

\subsection{Overall Architecture} 
With the primary objective of bolstering generalizability to unseen sources, we have devised a frequency domain learning network to effectively enhance deepfake detection capabilities. The comprehensive architecture of the FreqNet approach is depicted in Figure \ref{fig:overall}.
Within our methodology, we introduce practical and compact frequency learning plugin modules designed to compel the CNN classifier to operate within the frequency domain. These modules include the high-frequency representation and the frequency convolutional layer. The modular nature of these components allows seamless integration into the CNN classifier. This architectural innovation has culminated in the creation of a novel and lightweight detector named FreqNet, incorporating the frequency learning plugins alongside a limited number of CNN layers.

\subsubsection{High-Frequency Representation of Images} 
High-frequency artifacts have been recognized as valuable indicators for distinguishing between real and fake images \cite{qian2020thinking, luo2021generalizing}. Consequently, we leverage this valuable insight by adopting the high-frequency components of images as the input for the detector in our approach. 
In the case of each training image denoted as $x \in \mathbb{R}^{W \times H \times 3}$, our initial step involves converting it into the frequency domain using the Fast Fourier Transform (FFT). Subsequently, we proceed to extract the high-frequency components, represented as $f_h$, through the application of a high-pass filter denoted as $\mathcal{B}_h$:
\begin{equation}
 f_h = \mathcal{B}_h(\mathcal{F}(x)),
  \label{eq:eq4}
\end{equation}
where $\mathcal{F}$ denotes FFT, $f_h \in \mathbb{R}^{W \times H \times 3}$  denotes the frequency repersentation of images. The zero-frequency is shifted to the center. The high-pass filter $\mathcal{B}_h(\cdot)$ can be defined by:
\begin{equation}
\mathcal{B}_h(f_{i,j})=\left\{
\begin{aligned}
& f_{i,j}, otherwise, \\
& 0, if \left| i \right| < W/4, \left| j \right| < H/4
\end{aligned}
\right.
  \label{eq:eq4}
\end{equation}
where the center of the image is adopted as the origin.

Following the extraction of high-frequency components, we proceed to transform this frequency information back to image space:
\begin{equation}
 x_h =\mathcal{IF}(f_h),
  \label{eq:eq54}
\end{equation}
where the $\mathcal{IF}$ denotes inverse Fast Fourier Transform (iFFT). The result of this transformation, denoted as $x_h$, represents the high-frequency components within the image space. This process ensures that our focus remains on the pertinent high-frequency information contained within the image.

\subsubsection{High-Frequency Representation of Feature}
Indeed, the varying GAN architectures incorporate distinct frequency patterns, which can potentially lead the detector to overfit to the specifics of the training source. To mitigate this issue and enhance the detector's generalization capacity, we adopt a strategy that involves compelling the detector to consistently prioritize and focus on high-frequency information within the feature space. By emphasizing the importance of high-frequency cues in the feature space, we effectively counteract the overfitting tendencies, promoting a more robust and adaptable deepfake detection capability.

Specifically, for output of the $k_{th}$ convolutional layer denoted as $M^k \in \mathbb{R}^{H \times W \times C}$, we transform it to frequency space across spatial $(W,H)$ and  the channel dimension $C$ using the Fast Fourier Transform (FFT), respectively. The zero frequency component is moved to the center. Subsequently, we apply a high-pass filter $\mathcal{B}_h$ to extract the high-frequency information from this transformed representation. Finally, we reverse this frequency transformation, converting the extracted high-frequency information back to the feature space:
\begin{equation}
M^k_h(dim)=\left\{
\begin{aligned}
\mathcal{IF}_{W,H}(\mathcal{B}_h(\mathcal{F}_{W,H}(M^k))),&if dim=(W, H)\\
\mathcal{IF}_{C}(\mathcal{B}_h(\mathcal{F}_{C}(M^k))),        &if dim=C  
\end{aligned}
\right.
  \label{eq:eq54}
\end{equation}
where $M^k_h(W,H)$,  $M^k_h(C)$ represent the high-frequency components of feature maps $M^k$ acorss spatial $(W,H)$ and channel $C$ dimension in feature space, respectively. 
We implement two distinct high-frequency component extractors, each targeting different CNN layers within our method. This strategic differentiation allows us to leverage the unique information present at different stages of the network, enhancing the overall sensitivity to high-frequency cues.

\subsubsection{Frequency Convolutional Layer}
Many existing frequency-based approaches follow a paradigm of extracting frequency information from images and employing it to train a CNN classifier. However, this approach often results in the detector overfitting to the specifics of the training source, leading to suboptimal performance when faced with previously unseen sources.  In contrast, our approach not only employs the frequency information as the artifact representation. We also introduce frequency space learning as a strategy to significantly enhance the generalization ability of the detector.

Specifically, within our approach, the feature maps from the convolutional layers are initially transformed from the feature space to the frequency domain. Following this transformation, we apply convolutional layers on both the phase spectrum and the amplitude spectrum, thereby enabling the detector to learn within the frequency space. Subsequently, the learned spectrum information is transformed back to the feature space using the inverse Fast Fourier Transform. This comprehensive process effectively emphasizes the representation ability of the detector within the frequency domain, enhancing its sensitivity to critical features present in this space.

Let's consider the given feature maps $M^k \in \mathbb{R}^{W \times H \times C}$ from $k_{th}$ CNN layer. The learning process within the frequency space can be formally defined as follows:

\begin{equation}
\begin{aligned}
&f = f_{am} + f_{ph}\mathrm{i} = \mathcal{F}_{W,H}(M^k)\\
&\widetilde{f_{am}} = L_{conv}(f_{am}) \\
&\widetilde{f_{ph}} = L_{conv}(f_{ph}) \\
&\widetilde{M^k}   = \mathcal{IF}_{W,H}(\widetilde{f_{am}} + \widetilde{f_{ph}}\mathrm{i}) \\
\end{aligned}
  \label{eq:eq541}
\end{equation}
where $\mathcal{F}_{W,H}, \mathcal{IF}_{W,H}$ denote FFT and iFFT, $f_{am}, f_{ph}$ are amplitude spectrum and phase spectrum of feature maps $M^k$, and $L_{conv}$ denotes a CNN layer, $\widetilde{f_{am}}, \widetilde{f_{ph}}$ are the learned amplitude spectrum and phase spectrum of feature maps $M^k$. Subsequently, The feature map $\widetilde{M^k}$ learned in spectrum space can be calculated by iFFT.

As a result, the comprehensive training procedure for FreqNet can be formally defined as:
\begin{equation} 
D_{freq} = \mathop{\arg\min}_{\theta} \ l(D_{freq}( x_h; \theta),\  y),
  \label{eq:eq542}
\end{equation}
where $l()$ denotes the standard cross entropy loss, and $D_{freq}$ is our frequency sapce learning network. 
Our FreqNet harnesses spectrum learning to accomplish domain-invariant deepfake detection. Within our approach, we have meticulously designed two key modules: the high-frequency representation module and the frequency convolutional layer. 
Importantly, this work places emphasis on training our detector using a constrained amount of training data, followed by comprehensive evaluations in the challenging wild scenes, encompassing a diverse set of 17 GAN models.

\section{Experiments}
In this section, we provide a comprehensive evaluation of the FreqNet. We cover various aspects, including datasets, implementation details, deepfake detection performance, and more details to be described. Further elaborations on each of these aspects will be presented to offer a comprehensive understanding of the capabilities and effectiveness of FreqNet.

\begin{table*}[!ht]
    \centering
\resizebox{\textwidth}{45mm}{
    \begin{tabular}{l c c c c c c c c c c c c c c c c c c| c c}
    \bottomrule \hline
      \multirow{3}*{Methods} & \multicolumn{2}{c}{Settings}&\multicolumn{18}{c}{ Test Models}\\ 
       \cline{2-21} ~ & \multirow{2}*{Input}& \multirow{2}*{\#n} & \multicolumn{2}{c}{ProGAN}& \multicolumn{2}{c}{StyleGAN}& \multicolumn{2}{c}{StyleGAN2}& \multicolumn{2}{c}{BigGAN}& \multicolumn{2}{c}{CycleGAN}& \multicolumn{2}{c}{StarGAN}& \multicolumn{2}{c}{GauGAN}& \multicolumn{2}{c|}{Deepfake}& \multicolumn{2}{c}{Mean}\\
         \cline{4-21} ~ &~         & ~   & Acc. & A.P. & Acc. & A.P. & Acc. & A.P. & Acc. & A.P. & Acc. & A.P. & Acc. & A.P. & Acc. & A.P. & Acc. & A.P. & Acc. & A.P. \\ \bottomrule \hline
        Wang\shortcite{wang2020cnn}            & Img     & 1 & 50.4 & 63.8 & 50.4 & 79.3 & 68.2 & 94.7 & 50.2 & 61.3 & 50.0 & 52.9 & 50.0 & 48.2 & 50.3 & 67.6 & 50.1 & 51.5 & 52.5 & 64.9 \\ 
        Frank\shortcite{Frank}                          &  Freq       & 1 & 78.9 & 77.9 & 69.4 & 64.8 & 67.4 & 64.0 & 62.3 & 58.6 & 67.4 & 65.4 & 60.5 & 59.5 & 67.5 & 69.1 & 52.4 & 47.3 & 65.7 & 63.3 \\ 
        Durall\shortcite{Durall}                          &  Freq       & 1 & 85.1 & 79.5 & 59.2 & 55.2 & 70.4 & 63.8 & 57.0 & 53.9 & 66.7 & 61.4 & 99.8 & 99.6 & 58.7 & 54.8 & 53.0 & 51.9 & 68.7 & 65.0 \\ 
        F3Net\shortcite{qian2020thinking}&  Freq & 1  & 96.9 & 99.9 & 86.3 & 99.8 & 80.5 & 99.8 & 66.6 & 72.2 & 76.7 & 84.0 & 99.1 & 100.0 & 59.1 & 60.6 & 61.2 & 82.3 & 78.3 & 87.3 \\
        BiHPF\shortcite{jeong2022bihpf}        &  Freq       & 1 & 82.5 & 81.4 & 68.0 & 62.8 & 68.8 & 63.6 & 67.0 & 62.5 & 75.5 & 74.2 & 90.1 & 90.1 & {73.6} & {92.1} & 51.6 & 49.9 & 72.1 & 72.1 \\ 
        FrePGAN\shortcite{jeong2022frepgan}  & Img     & 1 & 95.5 & 99.4 & 80.6 & 90.6 & 77.4 & 93.0 & 63.5 & 60.5 & 59.4 & 59.9 & 99.6 & {100.0} & 53.0 & 49.1 & {70.4} & {81.5} & 74.9 & 79.3 \\ 
        LGrad \shortcite{Tan2023CVPR}          & Grad     & 1 & 99.4 & 99.9 & 96.0 & 99.6 & 93.8 & 99.4 & 79.5 & 88.9 & 84.7 & 94.4 & 99.5 & 100.0 & 70.9 & 81.8 & 66.7 & 77.9 & 86.3 & 92.7 \\          
         Ojha \shortcite{ojha2023towards}     & Fea     & 1 & 99.1 & 100.0 & 77.2 & 95.9 & 69.8 & 95.8 & 94.5 & 99.0 & 97.1 & 99.9 & 98.0 & 100.0 & 95.7 & 100.0 & 82.4 & 91.7 & \underline{89.2} & \tft{97.8} \\
        FreqNet              & Freq & 1  & 98.0 & 99.9 & 92.0 & 98.7 & 89.5 & 97.9 & 85.5 & 93.1 & 96.1 & 99.1 & 94.2 & 98.4 & 91.8 & 99.6 & 69.8 & 94.4 & \tft{89.6} & \underline{97.6}\\
\hline
        Wang\shortcite{wang2020cnn}   & Img & 2 & 64.6 & 92.7 & 52.8 & 82.8 & 75.7 & 96.6 & 51.6 & 70.5 & 58.6 & 81.5 & 51.2 & 74.3 & 53.6 & 86.6 & 50.6 & 51.5 & 57.3 & 79.6 \\ 
        Frank\shortcite{Frank}   &  Freq & 2 & 85.7 & 81.3 & 73.1 & 68.5 & 75.0 & 70.9 & 76.9 & 70.8 & {86.5} & 80.8 & 85.0 & 77.0 & 67.3 & 65.3 & 50.1 & 55.3 & 75.0 & 71.2 \\ 
        Durall\shortcite{Durall}  &  Freq & 2 & 79.0 & 73.9 & 63.6 & 58.8 & 67.3 & 62.1 & 69.5 & 62.9 & 65.4 & 60.8 & 99.4 & 99.4 & 67.0 & 63.0 & 50.5 & 50.2 & 70.2  & 66.4\\
        F3Net\shortcite{qian2020thinking} &  Freq & 2  & 97.9 & 100.0 & 84.5 & 99.5 & 82.2 & 99.8 & 65.5 & 73.4 & 81.2 & 89.7 & 100.0 & 100.0 & 57.0 & 59.2 & 59.9 & 83.0 & 78.5 & 88.1 \\
        BiHPF\shortcite{jeong2022bihpf}   &  Freq & 2 & 87.4 & 87.4 & 71.6 & 74.1 & 77.0 & 81.1 & {82.6} & 80.6 & 86.0 & 86.6 & 93.8 & 80.8 & \tf{75.3} & {88.2} & 53.7 & 54.0 & 78.4 & 79.1 \\
        FrePGAN\shortcite{jeong2022frepgan} & Img & 2 & 99.0 & 99.9 & 80.8 & 92.0 & 72.2 & 94.0 & 66.0 & 61.8 & 69.1 & 70.3 & 98.5 & {100.0} & 53.1 & 51.0 & {62.2} & {80.6} & 75.1 & 81.2 \\ 
        LGrad \shortcite{Tan2023CVPR}    & Grad    & 2 & 99.8 & 100.0 & 94.8 & 99.7 & 92.4 & 99.6 & 82.5 & 92.4 & 85.9 & 94.7 & 99.7 & 99.9 & 73.7 & 83.2 & 60.6 & 67.8 & 86.2 & 92.2 \\
         Ojha \shortcite{ojha2023towards}     & Fea     & 2 & 99.7 & 100.0 & 78.8 & 97.4 & 75.4 & 96.7 & 91.2 & 99.0 & 91.9 & 99.8 & 96.3 & 99.9 & 91.9 & 100.0 & 80.0 & 89.4 & \underline{88.1} & \underline{97.8} \\
        FreqNet              &  Freq & 2 &          99.6 & 100.0 & 90.4 & 98.9 & 85.8 & 98.1 & 89.0 & 96.0 & 96.7 & 99.8 & 97.5 & 100.0 & 88.0 & 98.8 & 80.7 & 92.0 & \tft{91.0} & \tft{97.9} \\
\hline
        Wang\shortcite{wang2020cnn}  & Img & 4 & 91.4 & 99.4 & 63.8 & 91.4 & 76.4 & 97.5 & 52.9 & 73.3 & 72.7 & 88.6 & 63.8 & 90.8 & 63.9 & \tf{92.2} & 51.7 & 62.3 & 67.1  &86.9 \\ 
High-Freq&  Freq &4 & 98.9 & 100.0 & 74.4 & 98.3  & 68.8 & 97.3 & 75.2 & 92.1 & 71.0 & 87.9 & 92.7 & 100.0 & 75.5 & 86.5 & 57.0 & 74.9 & 76.7 & 92.1 \\
        Frank\shortcite{Frank}  &  Freq & 4 & 90.3 & 85.2 & 74.5 & 72.0 & 73.1 & 71.4 & {88.7} & \tf{86.0} & 75.5 & 71.2 & 99.5 & 99.5 & 69.2 & 77.4 & 60.7 & 49.1 & 78.9 & 76.5 \\ 
        Durall\shortcite{Durall} &  Freq & 4 & 81.1 & 74.4 & 54.4 & 52.6 & 66.8 & 62.0 & 60.1 & 56.3 & 69.0 & 64.0 & 98.1 & 98.1 & 61.9 & 57.4 & 50.2 & 50.0 & 67.7  & 64.4\\ 
        F3Net\shortcite{qian2020thinking} &  Freq &4  & 99.4 & 100.0 & 92.6 & 99.7 & 88.0 & 99.8 & 65.3 & 69.9 & 76.4 & 84.3 & 100.0 & 100.0 & 58.1 & 56.7 & 63.5 & 78.8 & 80.4 & 86.2 \\
        BiHPF\shortcite{jeong2022bihpf}   &  Freq & 4 & 90.7 & 86.2 & 76.9 & 75.1 & 76.2 & 74.7 & 84.9 & 81.7 & 81.9 & 78.9 & 94.4 & 94.4 & 69.5 & 78.1 & 54.4 & 54.6 & 78.6  & 77.9\\
        FrePGAN\shortcite{jeong2022frepgan} & Img & 4 & 99.0 & 99.9 & 80.7 & 89.6 & 84.1 & 98.6 & 69.2 & 71.1 & 71.1 & 74.4 & {99.9} & {100.0} & 60.3 & 71.7 & {70.9} & {91.9} & 79.4  & 87.2\\ 
        LGrad \shortcite{Tan2023CVPR}    & Grad    & 4 & 99.9 & 100.0 & 94.8 & 99.9 & 96.0 & 99.9 & 82.9 & 90.7 & 85.3 & 94.0 & 99.6 & 100.0 & 72.4 & 79.3 & 58.0 & 67.9 & 86.1 & 91.5 \\
        Ojha \shortcite{ojha2023towards}     & Fea     & 4 & 99.7 & 100.0 & 89.0 & 98.7 & 83.9 & 98.4 & 90.5 & 99.1 & 87.9 & 99.8 & 91.4 & 100.0 & 89.9 & 100.0 & 80.2 & 90.2 & \underline{89.1} & \underline{98.3} \\
        FreqNet              &  Freq & 4 &          99.6 & 100.0 & 90.2 & 99.7 & 88.0 & 99.5 & 90.5 & 96.0 & 95.8 & 99.6 & 85.7 & 99.8 & 93.4 & 98.6 & 88.9 & 94.4 & \tft{91.5} & \tft{98.5}\\
\bottomrule
    \end{tabular}
}
  \caption{{Cross-model performance on the test set of ForenSynths\cite{wang2020cnn}.} \textbf{Bold} and \underline{underline} represent the best and second-best performance, respectively. }
  \label{tab:SOTA}
\end{table*}

\begin{table*}[!ht]
    \centering
\resizebox{\textwidth}{11.7mm}{
    \begin{tabular}{l  c c c c c c c c c c c c c c c c c c| c c}
    \bottomrule \hline
       \multirow{2}*{Method} & \multicolumn{2}{c}{AttGAN}& \multicolumn{2}{c}{BEGAN}& \multicolumn{2}{c}{CramerGAN}& \multicolumn{2}{c}{InfoMaxGAN}& \multicolumn{2}{c}{MMDGAN}& \multicolumn{2}{c}{RelGAN}& \multicolumn{2}{c}{S3GAN}& \multicolumn{2}{c}{SNGAN}&  \multicolumn{2}{c|}{STGAN}& \multicolumn{2}{c}{Mean}\\
         \cline{2-21} ~   & Acc. & A.P. & Acc. & A.P. & Acc. & A.P. & Acc. & A.P. & Acc. & A.P. & Acc. & A.P. & Acc. & A.P. & Acc. & A.P. & Acc. & A.P. & Acc. & A.P.\\ \bottomrule \hline
Wang\shortcite{wang2020cnn}    & 51.1 & 83.7 & 50.2 & 44.9 & 81.5 & 97.5 & 71.1 & 94.7  & 72.9 & 94.4 & 53.3 & 82.1 & 55.2 & 66.1 & 62.7 & 90.4 & 63.0 & 92.7 & 62.3 & 82.9 \\
F3Net\shortcite{qian2020thinking}   & 85.2 & 94.8 & 87.1 & 97.5 & 89.5 & 99.8 & 67.1 & 83.1 & 73.7 & 99.6 & 98.8 & 100.0 & 65.4 & 70.0 & 51.6 & 93.6 & 60.3 & 99.9 & 75.4 & 93.1 \\
LGrad \shortcite{Tan2023CVPR}   & 68.6 & 93.8 & 69.9 & 89.2 & 50.3 & 54.0 & 71.1 & 82.0 & 57.5 & 67.3 & 89.1 & 99.1 & 78.5 & 86.0 & 78.0 & 87.4 & 54.8 & 68.0 & 68.6 & 80.8\\
Ojha \shortcite{ojha2023towards}  & 78.5 & 98.3 & 72.0 & 98.9 & 77.6 & 99.8 & 77.6 & 98.9 & 77.6 & 99.7 & 78.2 & 98.7 & 85.2 & 98.1 & 77.6 & 98.7 & 74.2 & 97.8 & \underline{77.6} & \tft{98.8}\\
FreqNet                                        & 89.8 & 98.8 & 98.8 & 100.0 & 95.2 & 98.2 & 94.5 & 97.3 & 95.2 & 98.2 & 100.0 & 100.0 & 88.3 & 94.3 & 85.4 & 90.5 & 98.8 & 100.0 & \tft{94.0} & \underline{97.5}\\
\bottomrule
    \end{tabular}
}
  \caption{{Cross-model performance on the self-synthesis dataset.}}
  \label{tab:9gan}
\end{table*}

\begin{table}[!ht]
    \centering
\resizebox{68mm}{10.0mm}{
    \begin{tabular}{l  c c}
    \bottomrule
 Methods                                           &  Parameters $\downarrow$   & mAcc. $\uparrow$ of 17 models  \\  \hline
F3Net\shortcite{qian2020thinking}          & 48.9 M                    & 77.8                             \\     
LGrad\shortcite{Tan2023CVPR}              & \underline{46.6 M}    & 76.8                             \\
Ojha\shortcite{ojha2023towards}           & 304.0 M                   & \underline{83.0}                           \\
FreqNet                                                & \tft{1.9 M}                & \tft{92.8(+9.8)}                  \\     
\bottomrule
    \end{tabular} }
  \caption{{Comparison of Parameters.}  }
  \label{tab:Param}
\end{table}

\label{ER}
\subsection{Datasets}
\subsubsection{Training set.} To ensure a consistent basis for comparison, we employ the training set of ForenSynths \cite{wang2020cnn} to train the detectors, aligning with baselines \cite{wang2020cnn,jeong2022bihpf,jeong2022frepgan}. 
The training set consists of 20 distinct  categories, each comprising 18,000 synthetic images generated using ProGAN, alongside an equal number of real images sourced from the LSUN dataset. In line with previous research  \cite{jeong2022bihpf,jeong2022frepgan}, we adopt specific 1-class,  2-class, and 4-class training settings, denoted as (horse), (chair, horse), (car, cat, chair, horse), respectively.

\subsubsection{Real-world Scene Test set.}
To assess the generalization ability of the proposed method on the real-world scene, we adopt various images and diverse GAN models.  
Firstly, we employ the test set of ForenSynths\nocite{wang2020cnn} for evaluation. It includes fake images generated by 8 generation model~\footnote{ProGAN \cite{karras2018progressive}, StyleGAN \cite{karras2019style}, StyleGAN2 \cite{karras2020analyzing}, BigGAN \cite{BigGAN}, CycleGAN \cite{CycleGAN}, StarGAN \cite{choi2018stargan}, GauGAN \cite{GauGAN} and Deepfake \cite{Deepfake}}. The real images are sampled from 6 datasets~\footnote{LSUN \cite{yu2015lsun}, ImageNet \cite{russakovsky2015imagenet}, CelebA \cite{CelebA}, CelebA-HQ \cite{karras2018progressive}, COCO \cite{coco}, and FaceForensics++ \cite{Deepfake}}. 
Additionally, to replicate the unpredictability of wild scenes, we extend our evaluation by collecting images generated by 9 additional GANs~\footnote{AttGAN\cite{AttGAN}, BEGAN\cite{began}, CramerGAN\cite{CramerGAN}, InfoMaxGAN\cite{InfoMaxGAN}, MMDGAN\cite{MMDGAN}, RelGAN\cite{RelGAN}, S3GAN\cite{S3GAN}, SNGAN\cite{SNGAN}, and STGAN\cite{STGAN}}. There are 36K test images, with equal numbers of real and fake images.

Simultaneously, we curate a dedicated face test set comprising 20,000 real images sourced from Celeba-HQ \cite{karras2018progressive}, and 60,000 fake face images  from ProGAN \cite{karras2018progressive}, StyleGAN \cite{karras2019style}, and StyleGAN2 \cite{karras2020analyzing}.

\subsection{Implementation Details} 
We design a lightweight CNN classifier, employing residual convolutional blocks without pretraining.   During the training process, we utilize the Adam optimizer  \cite{kingma2015adam} with an initial learning rate of \(2 \times 10^{-2}\). The batch size is set at 32, and we train the model for 100 epochs.  
A learning rate decay strategy is employed, reducing the learning rate by twenty percent after every ten epochs. 
Consistent with established baselines \cite{jeong2022bihpf,jeong2022frepgan}, we utilize the average precision score (A.P.) and accuracy (Acc.)  as the primary evaluation metrics to gauge the effectiveness of our proposed method. These metrics provide a comprehensive assessment of the performance of our approach against the baselines. We employ the PyTorch framework \cite{paszke2019pytorch} for the implementation of our method, utilizing the computational power of the Nvidia GeForce RTX 3090 GPU. For the critical task of Fast Fourier Transform (FFT), we leverage the $torch.fft.fftn$ function within the PyTorch library.

\begin{figure}[t]
  \centering
   \includegraphics[scale=0.3]{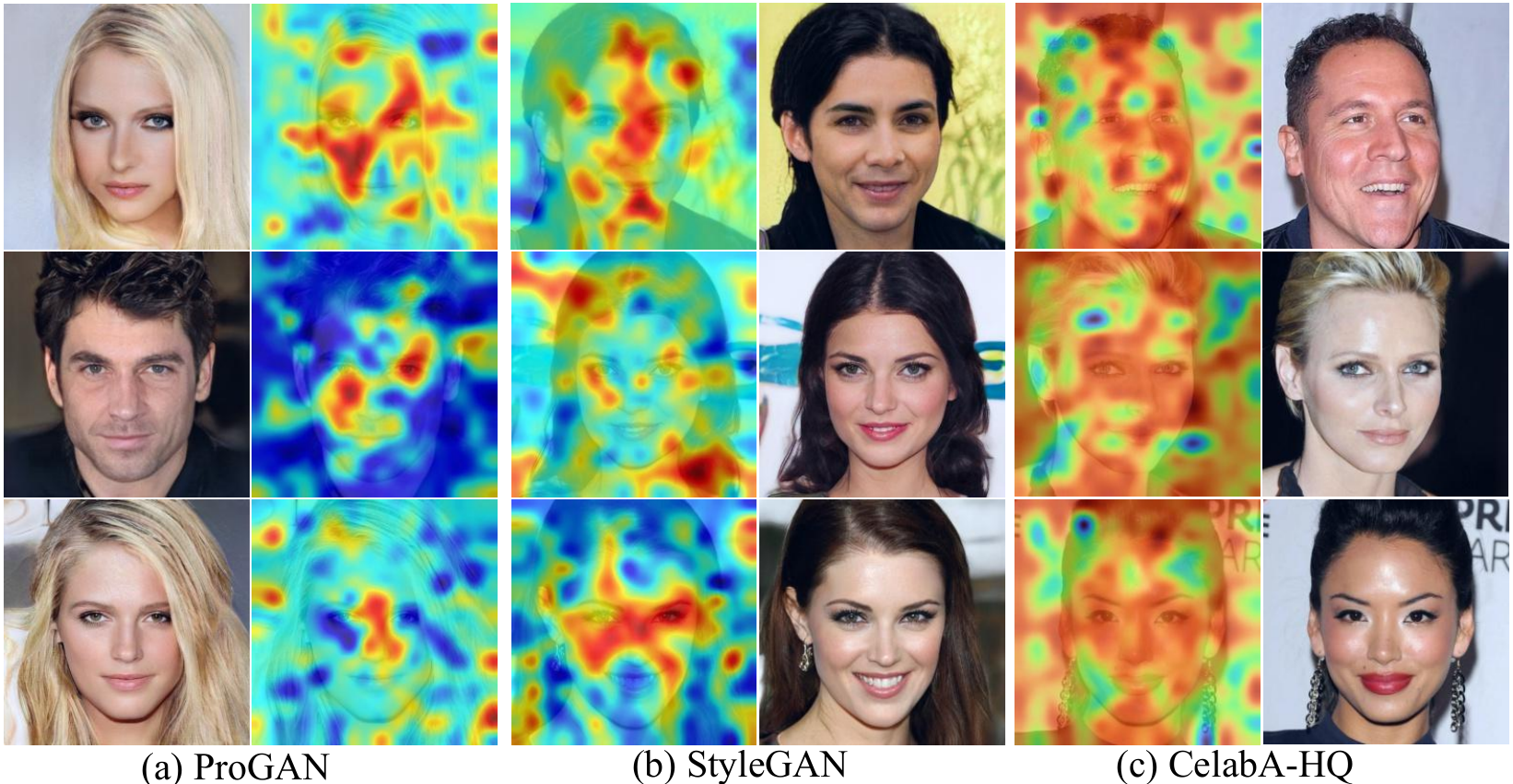}
   \caption{The visualization of Class Activate Map (CAM) \cite{zhou2016learning} extracted from detector on face images.}
   \label{fig:cam}
\end{figure}

\subsection{Deepfake Performance on Real-world Scene}

In order to demonstrate the remarkable generalization ability of our FreqNet on unseen sources,  we carry out evaluations on a real-world scene dataset. This dataset comprises images sourced from a total of 17 different generation models, encompassing 8 models from the ForenSynths test set and an additional 9 models from our own self-synthesis process.  This evaluation setup introduces increased complexity compared to the previous experiments, as the testing sets encompass a diverse range of GANs. This challenging scenario effectively simulates an open-world scene, making it a rigorous test of our approach's adaptability and robustness in real-world, unpredictable settings.

We compare with the previous methods: BiHPF \cite{jeong2022bihpf}, FreGAN \cite{jeong2022frepgan}, LGrad \cite{Tan2023CVPR}, Ojha \cite{ojha2023towards}. 
To ensure fair and meaningful comparisons, we adopt the same experimental setting as the established baselines \cite{jeong2022bihpf, jeong2022frepgan}. Specifically, the 1-class, 2-class and 4-class settings refer to training with the \textit{(horse)}, \textit{(chair, horse)}, \textit{(car, cat, chair, horse)} categories of ProGAN, respectively. 

The results of the ForenSynths dataset are presented in Table \ref{tab:SOTA}. 
The proposed FreqNet surpasses its counterparts in terms of mean Acc. metric and mean A.P. metrics, except for the value of A.P. on the 1-class setting.
Notably, with the 4-class setting, FreqNet achieves a mean Acc. value of 91.5\%, demonstrating its strong performance. 
In comparison to the current state-of-the-art methods LGrad and Ojha, our FreqNet exhibits substantial improvements, surpassing these methods by 5.4\% and 2.4\% in mean Acc., using fewer parameters.
 In the 1- and 2-class settings, our FreqNet also achieves gains of 2.9\% and 0.4\% compared to Ojha. 
Furthermore, compared to FingerprintNet\cite{jeong2022fingerprintnet} tested on six unseen models, our FreqNet achieves a marked improvement in mean accuracy, rising from 82.6\% to 90.6\% and exhibiting a significant gain of 8.0\%. 
Additionally, we provide results on 9 models from self-synthesis in Table \ref{tab:9gan}. We adopt the 4-class setting detector to perform testing. Compared to Ojha, our FreqNet achieves a marked improvement in mean accuracy, soaring from 77.6\% to an impressive 94.0\%, resulting in a significant gain of 16.4\%. When testing on face images, the proposed FreqNet  achieves accuracy rates of 98.7\%, 99.0\%, and 99.5\% on the ProGAN, StyleGAN, and StyleGAN2 datasets, respectively.

Furthermore, we provide a comprehensive overview of the number of parameters and the mean accuracy across all real-world scenes in Table \ref{tab:Param}.   It is evident that our FreqNet, with a modest parameter count of 1.9 million, significantly outperforms the current state-of-the-art model Ojha \cite{ojha2023towards}, which boasts an extensive 304 million parameters. This substantial difference in parameter count translates into a notable performance gain, with our FreqNet achieving a remarkable improvement of 9.8\% in mean accuracy compared to the larger model. This result underscores the efficiency and effectiveness of our FreqNet approach, demonstrating that superior performance can be achieved with significantly fewer parameters, a crucial advantage in real-world applications.

Compared to other frequency-based methods, such as BiHPF \cite{jeong2022bihpf}, FrePGAN \cite{jeong2022frepgan}, F3Net\cite{qian2020thinking}, our FreqNet achieves better performance on the real-world scene. 
The results confirm the generalization capability of the proposed frequency domain learning to extract a general representation of artifacts, and generalize this representation across various GAN models and categories.

\begin{table}[!ht]
    \centering
    \begin{tabular}{c c c c | c}
    \bottomrule
HFRI       &      HFRFS    &      HFRFC    &       FCL    &  mean Acc.  \\  \hline
           &   \checkmark  &   \checkmark  &   \checkmark & 84.3        \\     
\checkmark &               &   \checkmark  &   \checkmark & 85.3        \\     
\checkmark &   \checkmark  &               &   \checkmark & 87.8        \\     
\checkmark &               &               &   \checkmark & 82.0        \\     
\checkmark &   \checkmark  &   \checkmark  &              & 83.8        \\     
\checkmark &   \checkmark  &   \checkmark  &   \checkmark & 91.5        \\     
\bottomrule
    \end{tabular} 
  \caption{{ Ablation Study of FreqNet on the ForenSynths\cite{wang2020cnn}.}  }
  \label{tab:Ablation}
\end{table}

We perform ablation analyses on our FreqNet by individually removing the proposed modules. The results of these ablation experiments are presented in Table \ref{tab:Ablation}. Upon removal of the designed modules, we observe a decline in the detection performance, underscoring the efficacy of the proposed components. In the revised version, we will expound further on the specifics of the ablation analysis to provide a more comprehensive understanding.

\textbf{Visualization of Class Activate Map.} To visually demonstrate the discriminative regions identified by our detector, we present the Class Activation Maps (CAM) in Figure \ref{fig:cam}. The CAMs are generated using images from ProGAN, StyleGAN, StyleGAN2, and CelebA-HQ. The CAMs provide insights into the areas of focus for the detector in distinguishing real from fake images. It's noteworthy that the CAMs for real images highlight a broader portion of the image, while the CAMs for fake images tend to emphasize localized regions. Interestingly, even though the detector is primarily trained using a dataset containing cars, cats, chairs, and horses, it showcases the ability to recognize face images effectively. This highlights the versatility and adaptability of our detector in identifying distinct deepfake characteristics, even beyond the classes it was primarily trained on.

\section{Conclusion}

This study has focused on the introduction of FreqNet, a lightweight frequency space learning network designed for the task of generalizable forgery image detection. Our approach capitalizes on the power of frequency domain learning, offering an adaptable solution for the challenging problem of deepfake detection across diverse sources and GAN models. 
Within our methodology, we introduce practical and compact frequency learning plugin modules designed to compel the CNN classifier to operate within the frequency domain. 
The extensive experiments conducted on 17 different generation models serve as compelling evidence of FreqNet's generalization ability. This research contributes to advancing the field of deepfake detection, showcasing the potential of FreqNet to effectively combat the challenges posed by evolving forgery techniques and diverse image sources.

\section{Acknowledgments}
This work was supported in part by the National Key R\&D Program of China (No.2021ZD0112100), National NSF of China (No.U1936212, No.62120106009), National Natural Science Foundation of China under Grants 62072394, Natural Science Foundation of Hebei province under Grant F2021203019, A*STAR Career Development Funding ({CDF}) Award (Grant No:{222D800031}) and the Fundamental Research Funds for the Central Universities+2022YJS026.

\bibliography{egbib}

\end{document}